\documentclass{article}
\usepackage{spconf}

\usepackage{amsmath, amsfonts, graphicx, booktabs, tabularx, multirow, siunitx, xcolor, soul, bm}
\usepackage{hyperref}
\graphicspath{{./figures/}}

\usepackage{enumitem}
\usepackage{url}
\setlist{nosep, leftmargin=14pt}

\usepackage{mwe} 

\begin{document}

\title{{Channel Scaling: A Scale-and-Select Approach for Transfer Learning}}

\name{Ken C. L. Wong, Satyananda Kashyap, Mehdi Moradi$^{*}$
\thanks{This paper was accepted by IEEE ISBI 2021. \copyright 2021 IEEE. Personal use of this material is permitted. Permission from IEEE must be obtained for all other uses, in any current or future media, including reprinting/republishing this material for advertising or promotional purposes, creating new collective works, for resale or redistribution to servers or lists, or reuse of any copyrighted component of this work in other works.
}
\thanks{
\noindent$^{*}$Corresponding author.
}
}
\address{IBM Research -- Almaden Research Center, San Jose, CA, USA \\
{\footnotesize
\texttt{clwong@us.ibm.com, satyananda.kashyap@ibm.com, mmoradi@us.ibm.com}
}
}

\maketitle              

\begin{abstract}
Transfer learning with pre-trained neural networks is a common strategy for training classifiers in medical image analysis. Without proper channel selections, this often results in unnecessarily large models that hinder deployment and explainability. In this paper, we propose a novel approach to efficiently build small and well performing networks by introducing the channel-scaling layers. A channel-scaling layer is attached to each frozen convolutional layer, with the trainable scaling weights inferring the importance of the corresponding feature channels. Unlike the fine-tuning approaches, we maintain the weights of the original channels and large datasets are not required. By imposing L1 regularization and thresholding on the scaling weights, this framework iteratively removes unnecessary feature channels from a pre-trained model. Using an ImageNet pre-trained VGG16 model, we demonstrate the capabilities of the proposed framework on classifying opacity from chest X-ray images. The results show that we can reduce the number of parameters by 95\% while delivering a superior performance.
\end{abstract}

\section{Introduction}


The area of automatic medical image interpretation has witnessed a major transformation with the increased popularity of deep learning. For medical image classification, in the absence of large training sets of annotated medical images, transfer learning which uses models pre-trained on large datasets such as the ImageNet \cite{Journal:Russakovsky:IJCV2015} is a common approach. If the size of the target dataset available is small, researchers can manually select and freeze some layers, usually the low-level ones, to provide pre-trained features for the subsequent trainable layers. If the size of the target dataset is large enough, some or all layers can be fine-tuned for better performance. In fact, as fine-tuning usually involves modifications of millions of parameters, it requires an amount of data which is usually unavailable in medical imaging. Furthermore, regardless of the more complicated procedures, the performance gained by fine-tuning can be limited especially on large datasets \cite{Conference:Raghu:NIPS2019}. As a result, transfer learning in medical imaging is usually used without fine-tuning and a bulk of pre-trained layers are used without detailed selections.


Although this type of transfer learning can simplify the training of new classifiers, there are two shortcomings. First, the wholesale use of large neural networks trained on the ImageNet most likely results in unnecessarily large models, and this is unfavourable for applications that run on the cloud or mobile devices. Secondly, the bulk of feature channels pre-trained on natural images without detailed selections may reduce the explainability in medical applications. The features driving the performance on medical images could be a subset of the thousands of feature channels within the network. Without removing the unnecessary feature channels, it can be difficult to perform channel-level investigations to understand the impact of each channel on the results. This creates an obstacle for the use of these classifiers in products and services that require regulatory reviews. In fact, the black-box nature of deep learning in general, and of transfer learning in particular, as a solution to avoid feature engineering can hinder the widespread use of artificial intelligence (AI) in radiology \cite{geis2019ethics}.


To select the appropriate channels for a specific problem, network pruning is used in the computer vision community \cite{Journal:Molchanov:arXiv2016,Journal:Cheng:arXiv2017,Journal:Frankle:arXiv2018}. Network pruning can reduce the numbers of parameters by 90\% without harming the network performance. In general, network pruning involves the iterations of three main steps: evaluating the importance of feature channels, removing less important channels, and fine-tuning. As mentioned above, fine-tuning requires a relatively large amount of data that may not be available in medical imaging. Moreover, iterations with fine-tuning gradually modified the weights of the feature channels, and this can reduce the reusability of the selected channels on similar problems. Another approach in adaptive transfer learning targets layer by layer tuning and selection using a policy network \cite{Conference:Guo:CVPR2019}.


In this paper, we propose a new approach for transfer learning that allows feature channel selections, without changing the values of the original weights of the network. The core idea is the introduction of the concept of a channel-scaling layer. The channel-scaling layer is added after each frozen convolutional layer to infer the importance of each channel, with the scaling weights trainable by backpropagation. This idea is originated by modeling the channel selection problem as a binary optimization problem in which the selection of a channel is indicated by a binary index. As this approach is computationally infeasible, we relax the binary constraint and add thresholding to enable selection. Using a target dataset, we train the scaling weights and remove channels by applying a threshold to the weights for multiple iterations. L1 regularization is also imposed on each channel-scaling layer to increase the sparsity of the learned model. To the best of our knowledge, the idea of introducing a channel-scaling layer for network pruning is novel. Using an ImageNet pre-trained VGG16 model \cite{Journal:Simonyan:arXiv2014}, we demonstrate the capabilities of the proposed framework on classifying opacity from chest X-ray images. The results show that we can reduce the number of parameters by 95\% while delivering a superior performance. When L1 regularization is used, our method delivers a network that only includes 969 channels compared to the original 4,224 in VGG16.

\section{Methodology}

\subsection{Problem Formulation}

As there are thousands of convolutional feature channels even for relatively small networks (e.g., 4,224 for VGG16), selecting the appropriate channels for transfer leaning is a difficult task. While the most common approach is selecting the layers with low-level features, this may include some unnecessary low-level channels and discard some useful high-level channels. For more comprehensive approaches of network pruning \cite{Journal:Molchanov:arXiv2016,Journal:Frankle:arXiv2018}, although they are promising on reducing the network size when maintaining the accuracy, the required fine-tuning process can be computationally expensive and data demanding. Alternatively, we can formulate the channel selection problem as an optimization problem. Let $c$ be the total number of convolutional channels of a pre-trained network, the goal of the optimization problem is to find a $c$-dimensional binary vector $\mathbf{s} \in \{0, 1\}^{c}$ (i.e., 2$^c$ combinations) that indicates which channels to be kept for optimal performance. Similar to neural architecture search \cite{Journal:Zoph:arXiv2016,Conference:Wong:MICCAI2019}, each iteration of updating $\mathbf{s}$ requires a network training, thus this approach is computationally infeasible as $c$ is usually large. To address these issues, as we observe that backpropagation is a very powerful tool for training millions of parameters, here we propose a framework around it for computationally feasible feature selection.

\subsection{Channel-Scaling Layer}

To select the channels of a pre-trained network with reduced computational complexity and data size requirement, here we introduce the simple but effective \emph{Channel-Scaling} layer. For the optimization problem mentioned above, by relaxing the requirement from $\mathbf{s} \in \{0, 1\}^{c}$ to $\mathbf{s} \in [0, 1]^{c}$, i.e., from binary to real numbers between 0 and 1, we can utilize backpropagation to obtain $\mathbf{s}$. A channel-scaling layer which takes input from a frozen (non-trainable) pre-trained convolutional layer $l$ with $c_l$ feature channels is given as:
\begin{align}
\label{eq:channel_scaling}
\hat{\mathbf{x}} = ChannelScaling(\mathbf{x}; \mathbf{s}_l) = [s_{l1} \mathbf{x}_1, \ldots, s_{lc_l} \mathbf{x}_{c_l}]
\end{align}
where $\mathbf{x}$ and $\hat{\mathbf{x}} \in \mathbb{R}^{h \times w \times c_l}$ are the input and output feature tensors, respectively, and $\mathbf{x}_{i} \in \mathbb{R}^{h \times w}$ contains the spatial features of channel $i$. $\mathbf{s}_l = (s_{l1}, \ldots, s_{lc_l}) \in [0, 1]^{c_l}$ comprises the scaling weights trainable by backpropagation, and each $\mathbf{x}_i$ is rescaled by $s_{li}$ to produce $\hat{\mathbf{x}}_i$. L1 regularization can also be imposed on each channel-scaling layer to increase the sparsity of the learned $\mathbf{s}_l$. Therefore, by training a network augmented by channel-scaling layers, each channel-scaling layer can learn the relative importance of the channels in the corresponding convolutional layer, and $\mathbf{s}$ can be obtained by concatenating all $\mathbf{s}_l$.

\begin{figure}[t]
    \centering
    \begin{minipage}[t]{1\linewidth}
      \centering
      \includegraphics[width=1\linewidth]{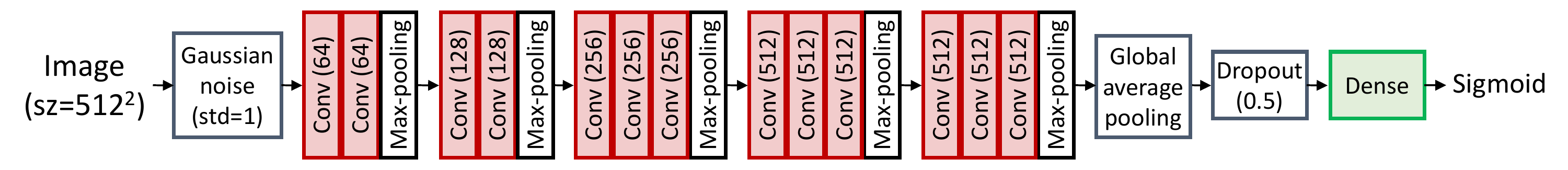}
      \centering{(a) Baseline architecture.}
    \end{minipage}
    \\
    \medskip
    \centering
    \begin{minipage}[t]{1\linewidth}
      \centering
      \includegraphics[width=1\linewidth]{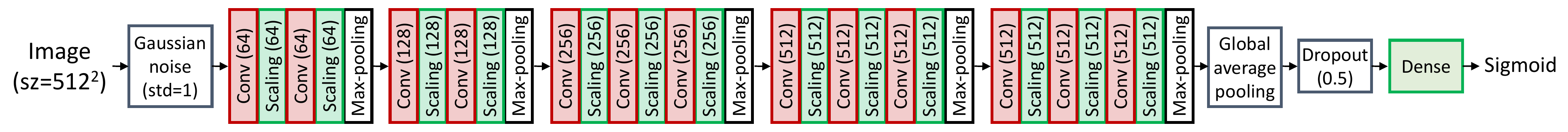}
      \centering{(b) Architecture with channel-scaling layers.}
    \end{minipage}
    \caption{Network architectures with ImageNet pre-trained convolutional layers from VGG16 (13 convolutional layers, 4,224 feature channels). Conv($c_l$) are 3$\times$3 convolutional layers with $c_l$ feature channels and ReLU. Scaling($c_l$) are the corresponding channel-scaling layers. Red blocks are non-trainable and green blocks are trainable.}
    \label{fig:networks}
\end{figure}

After adding the channel-scaling layers, the last convolutional features are pooled by global average pooling and a trainable final fully-connect layer is used to provide the predictions (Fig. \ref{fig:networks}). As the pre-trained convolutional layers are frozen, instead of millions of weights, only the $c$ scaling weights of the channel-scaling layers and the weights of the final fully-connected layer need to be trained.

Note that the channel-scaling layer is different from the squeeze-and-excitation block in \cite{Conference:Hu:CVPR2018} as the excitation scalars are input dependent. In contrast, the scaling weights in our channel-scaling layer are fixed after training.

\subsection{Scale-and-Select Strategy for Channel Selection}

After a network augmented with channel-scaling layers is trained, $\mathbf{s}_l$ can be used to remove the less important channels from the pre-trained convolutional layer $l$, for examples, those with the corresponding $s_{li} < 0.01$. Suppose the kernel shape of convolutional layer $l$ is ($k$, $k$, $c_{l-1}$, $c_l$), with $k$ the kernel size, and $c_{l-1}$ and $c_l$ are the numbers of input and output channels. If $n_{l-1}$ and $n_l$ channels are removed from the previous and this convolutional layers, respectively, the new kernel shape becomes ($k$, $k$, $c_{l-1} - n_{l-1}$, $c_l - n_l$). The corresponding layer biases are also removed accordingly. If all channels are removed from a layer, that layer and the subsequent convolutional layers are removed, though this did not happen in our experiments. Note that the remaining convolutional kernels are unaltered. After all layers are processed, we can attach the channel-scaling layers again and perform another iteration of training. This scale-and-select process iterates until some convergence criteria are fulfilled, for examples, reaching the maximum number of iterations or less than a number of channels can be removed.

To produce the final model after convergence, the feature channels are again removed according to $\mathbf{s}_l$, but this time the remaining kernel weights are multiplied by the corresponding $s_{li}$. No channel-scaling layers are added and only the final fully-connected layer is trained to obtain the final model.

Note that the selected channels, with or without scaling, can be further used with other transfer learning methods to produce models of better performance. These results cannot be presented because of the page limit.

\begin{figure}[t]
    \centering
    \begin{minipage}[t]{0.48\linewidth}
      \centering
      \includegraphics[width=1\linewidth]{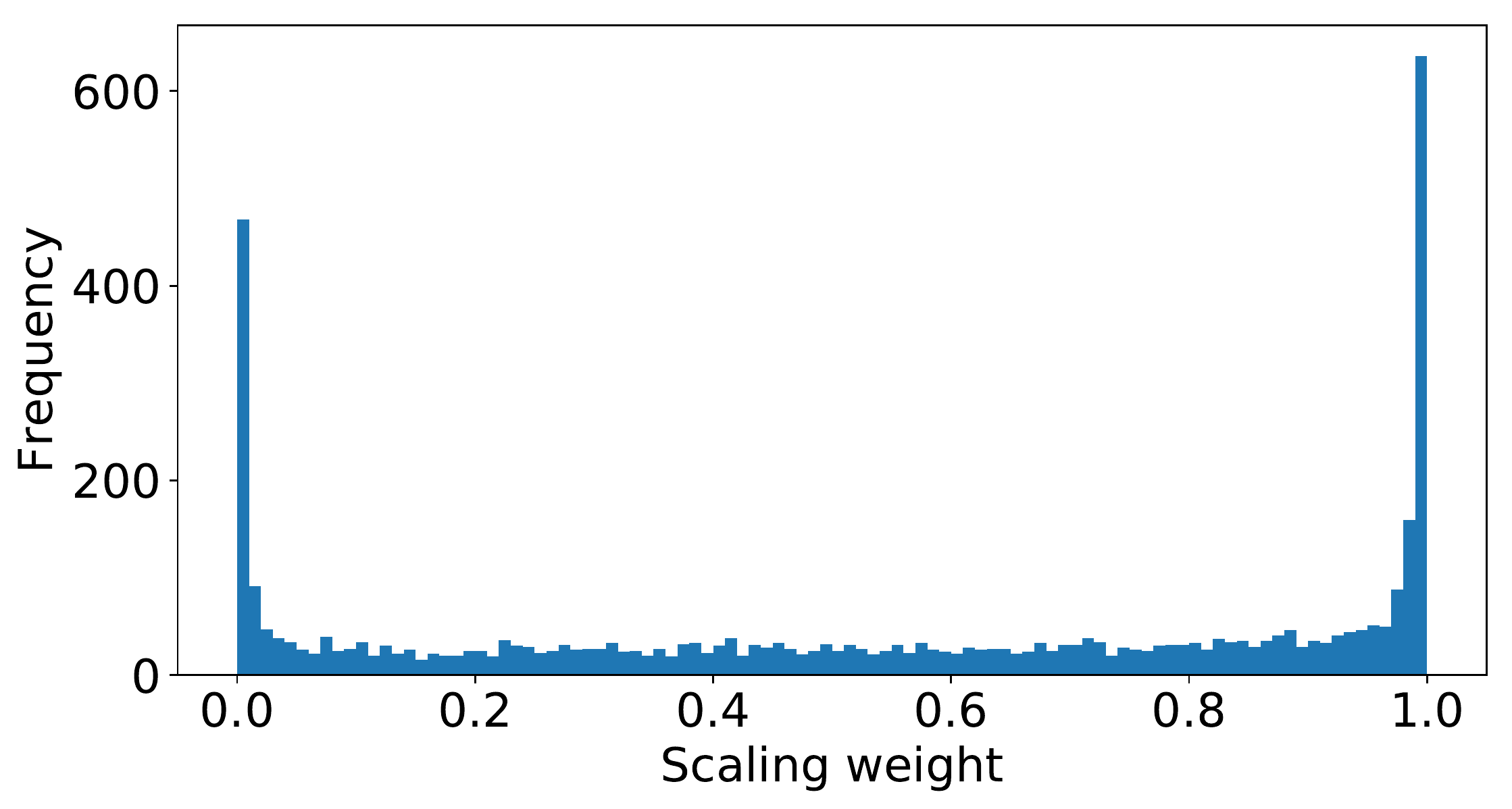}
      \centering{(a) No L1 regularization.}
    \end{minipage}
    \begin{minipage}[t]{0.48\linewidth}
      \centering
      \includegraphics[width=1\linewidth]{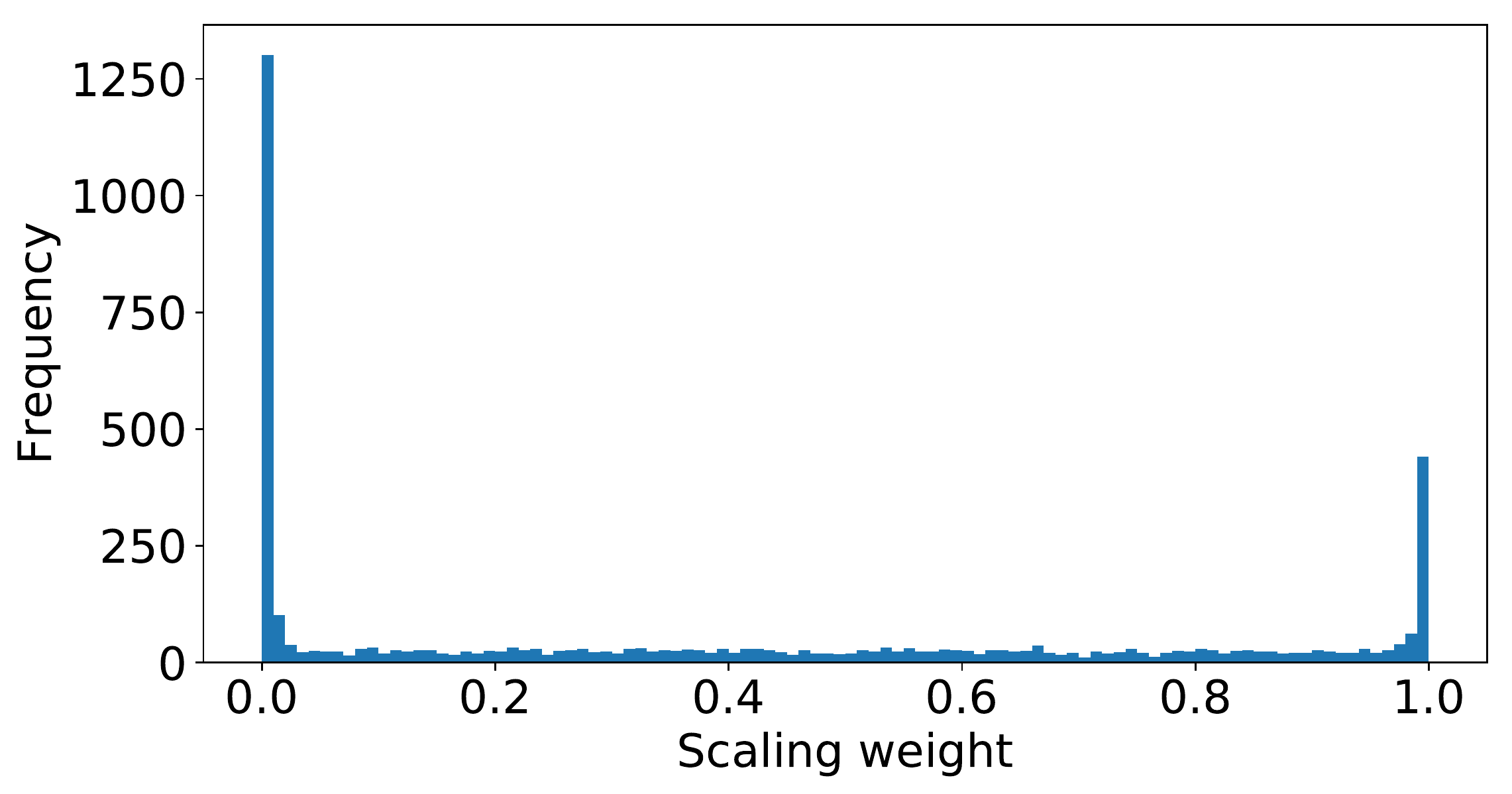}
      \centering{(b) With L1 regularization.}
    \end{minipage}
    \caption{Comparison of the scaling weight distributions between with and without L1 regularization after the first training. There are 100 bins between 0 and 1 and the total number of weights is 4,224.}
    \label{fig:plot_hist}
\end{figure}

\subsection{Training Strategy}

In each network training, image augmentation with rotation ($\pm$\ang{10}) and shifting ($\pm$10\%) is used with an 80\% chance. The optimizer Nadam is used with the learning rate of 5$\times$10$^{-4}$ and the binary crossentropy as the loss. Two NVIDIA Tesla V100 GPUs with 16 GB memory were used for multi-GPU training with a batch size of 32 and 50 epochs. In case L1 regularization is utilized, the regularization parameter of 10$^{-5}$ is used to balance between sparsity and accuracy.

\section{Experiments}

\subsection{Data and Experimental Setups}

We validated our framework on the MIMIC-CXR dataset \cite{Journal:Johnson:arXiv2019}. We focused on the binary classification of opacity, and 254,806 frontal images were used with 201,168 positive and 53,638 negative cases. The dataset was split into 20\% for training, 10\% for validation, and 70\% for testing in terms of patient ID with the positive to negative ratio maintained. Each image was resized to 512$\times$512.

As a proof of concept, the ImageNet pre-trained convolutional layers (13 layers, 4,224 feature channels) from VGG16 was used (Fig. \ref{fig:networks}). Please note that our goal is not to compete with the state-of-the-art classification performance but to study the characteristics of the proposed framework under different settings. The baseline model without channel-scaling layers was obtained by training only the final fully-connected layer (Fig. \ref{fig:networks}(a)). Two other models were trained using the scale-and-select strategy, but only one of them used the L1 regularization (regularization parameter = 10$^{-5}$) to increase the weights sparsity of the channel-scaling layers (Fig. \ref{fig:networks}(b)). In the channel selection process, the feature channels with the corresponding $s_{li} < 0.01$ were removed.

\subsection{Results and Discussion}

\begin{figure}[t]
    \centering
    \begin{minipage}[t]{0.47\linewidth}
      \centering
      \includegraphics[width=1\linewidth]{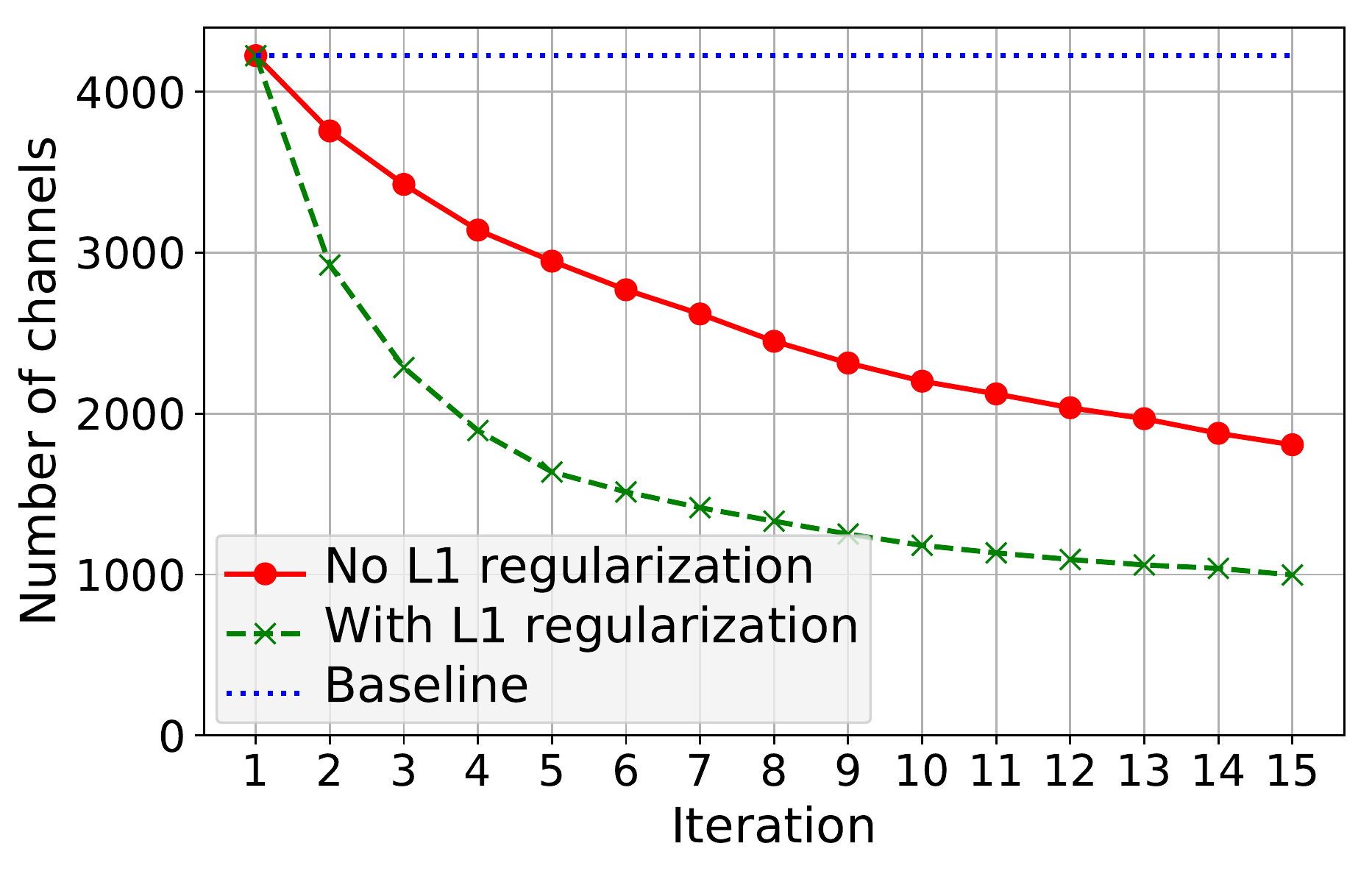}
      \centering{(a) Number of channels.}
    \end{minipage}
    \begin{minipage}[t]{0.47\linewidth}
      \centering
      \includegraphics[width=1\linewidth]{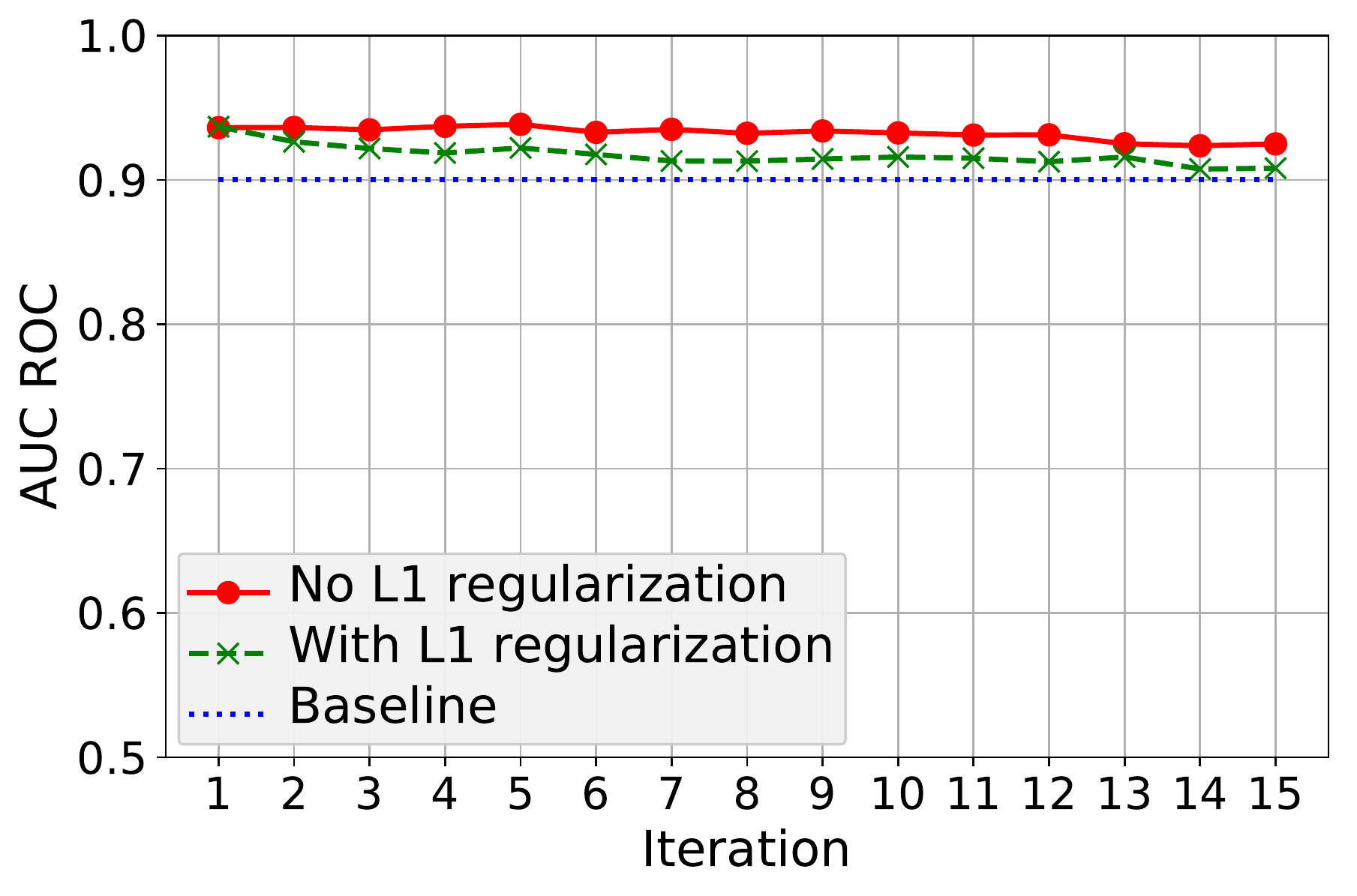}
      \centering{(b) AUC ROC.}
    \end{minipage}
    \\
    \medskip
    \begin{minipage}[t]{0.47\linewidth}
      \centering
      \includegraphics[width=1\linewidth]{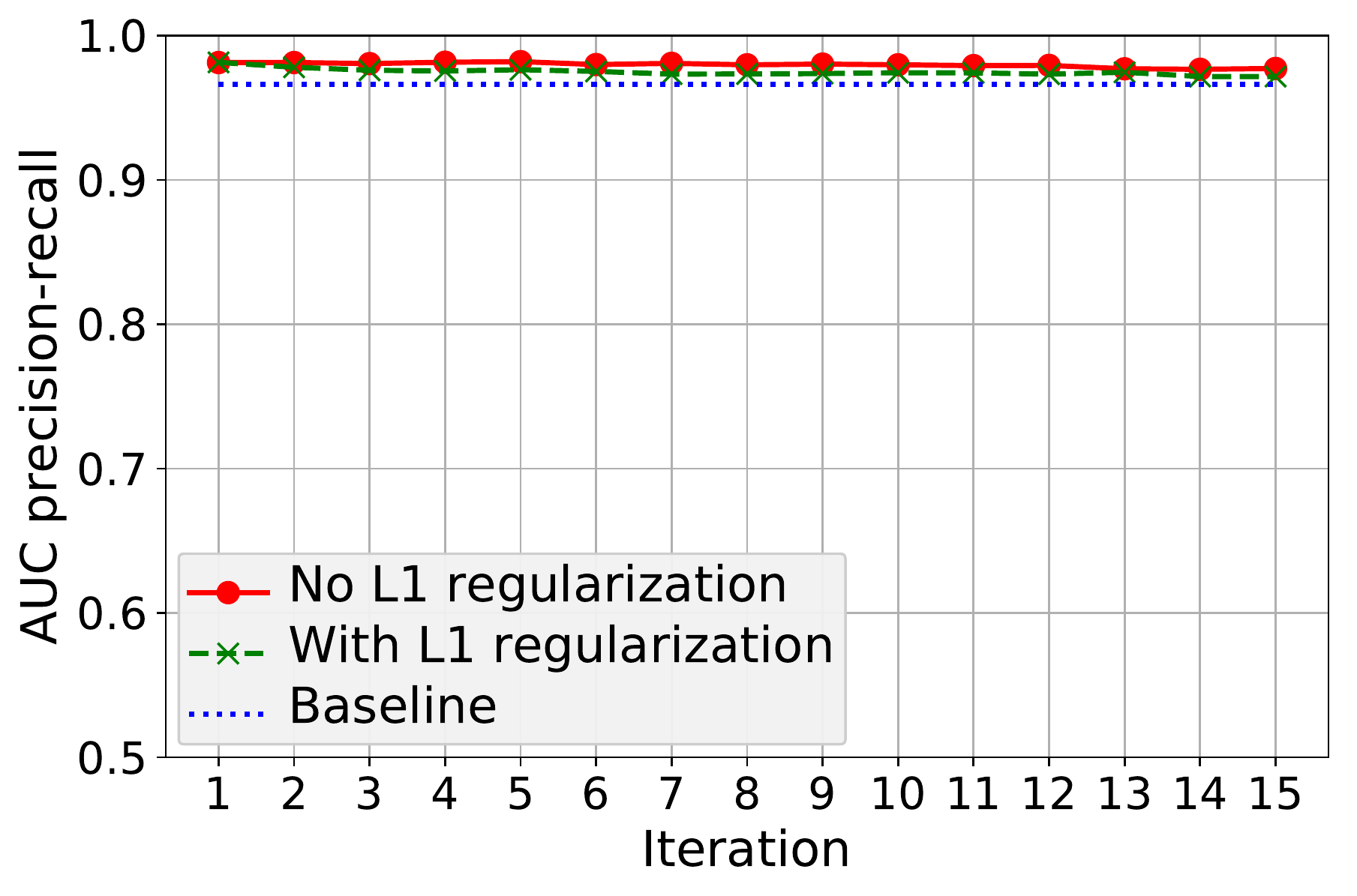}
      \centering{(c) AUC precision-recall.}
    \end{minipage}
    \caption{Scale-and-select iterations. AUC means area under curve. (a) Numbers of remaining channels. (b) and (c) The AUC of the ROC and precision-recall curves.}
    \label{fig:plot_itr}
\end{figure}

Fig. \ref{fig:plot_hist} shows the comparison of the scaling weight distributions between with and without L1 regularization after the first training. Without L1 regularization, there were 468 out of 4,224 scaling weights $<$ 0.01 and 636 weights $>$ 0.99. When L1 regularization is utilized, there were 1,301 weights $<$ 0.01 and 441 weights $>$ 0.99. Therefore, L1 regularization can lead to larger network size reduction.

Fig. \ref{fig:plot_itr} shows the evolution of the network properties with respect to the scale-and-select iterations. Both networks with and without L1 regularization had their numbers of channels decreasing with iterations, though the one with L1 regularization decreased faster. The rates of channel reduction slowed down with iteration and show a sign of convergence, especially for the one with L1 regularization. For the classification performance, both networks with channel-scaling layers performed better than the baseline even after 15 iterations, which is expectable as they had more trainable weights. There was slight performance penalty with respect to channel reduction without L1 regularization (AUC ROC: 0.936 to 0.925; AUC precision-recall: 0.981 to 0.977), while the penalty was more obvious with L1 regularization (AUC ROC: 0.937 to 0.909; AUC precision-recall: 0.981 to 0.972).

\begin{figure}[t]
    \centering
    \begin{minipage}[t]{0.32\linewidth}
      \centering
      \includegraphics[width=1\linewidth]{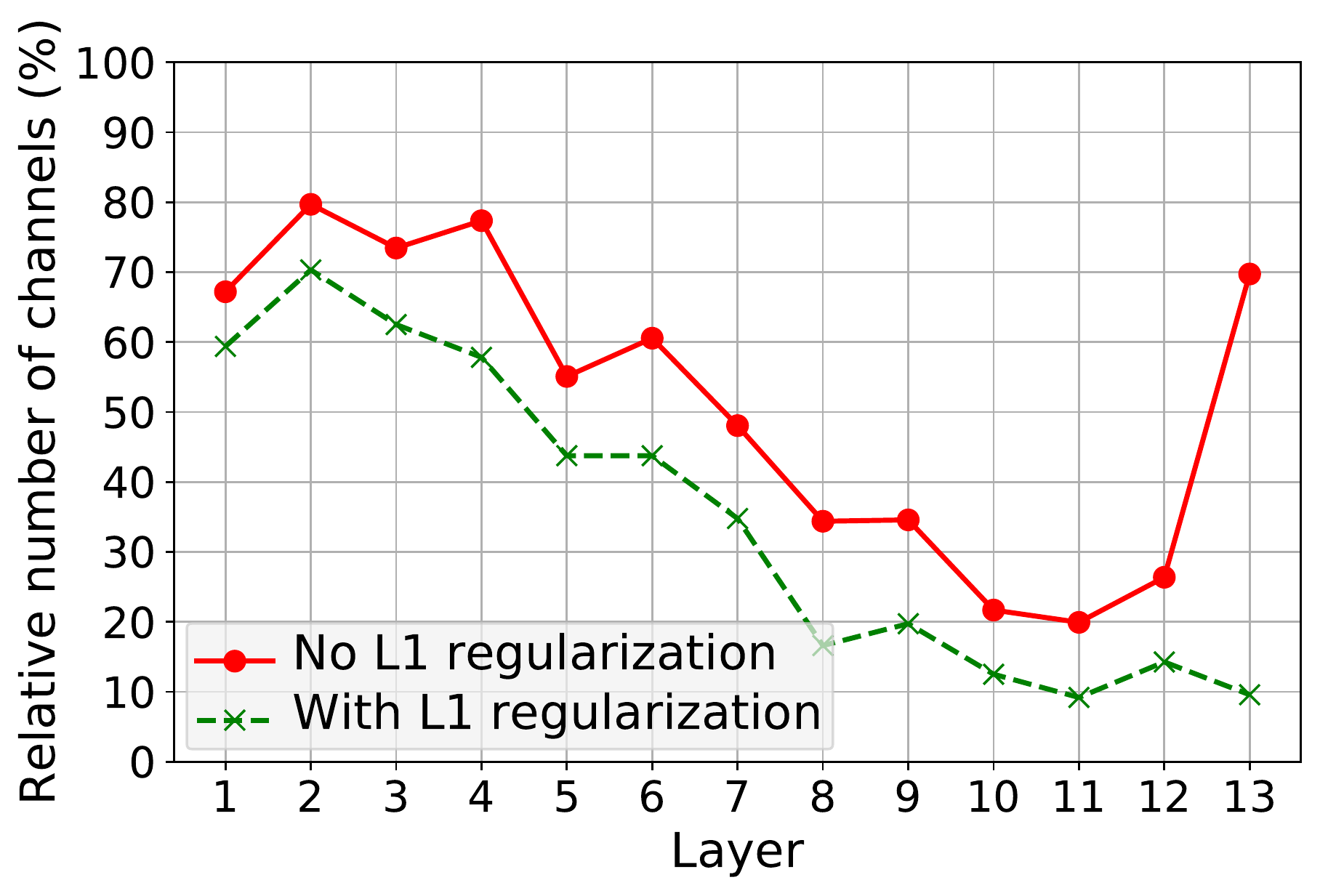}
      \centering{(a) Channels per layer.}
    \end{minipage}
    \begin{minipage}[t]{0.32\linewidth}
      \centering
      \includegraphics[width=1\linewidth]{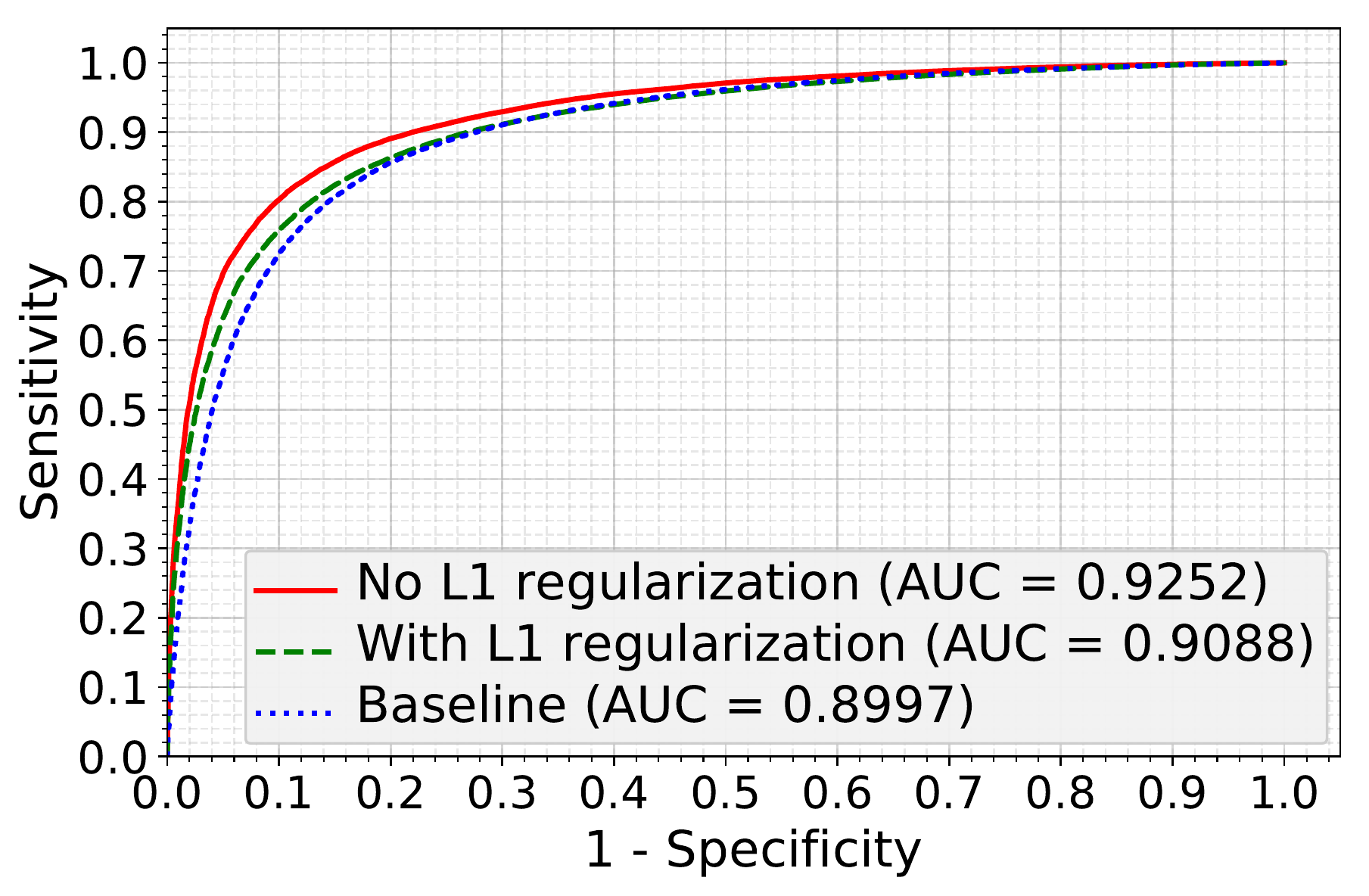}
      \centering{(b) ROC curve.}
    \end{minipage}
    \begin{minipage}[t]{0.32\linewidth}
      \centering
      \includegraphics[width=1\linewidth]{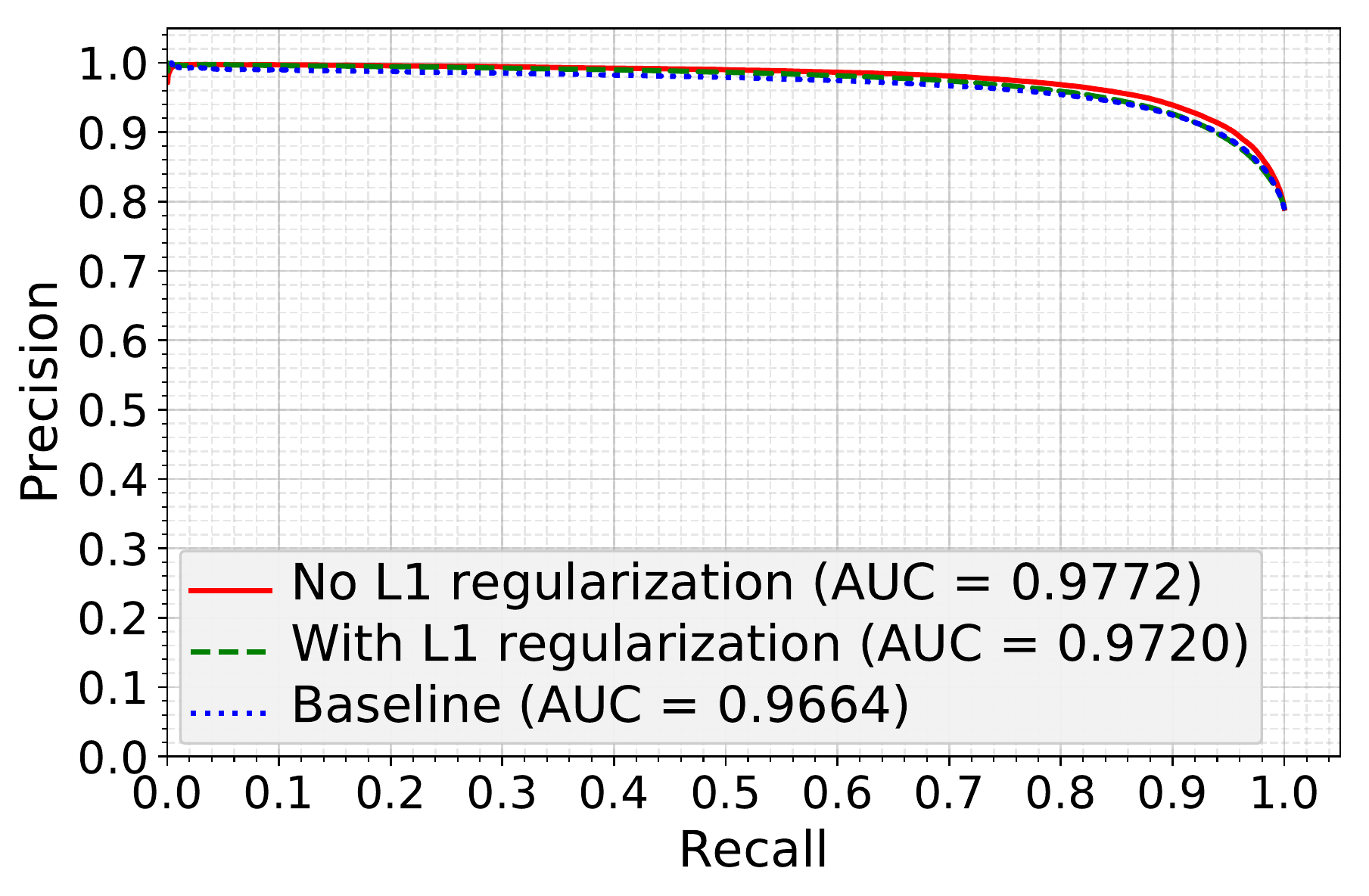}
      \centering{(c) Precision-recall curve.}
    \end{minipage}
    \caption{Properties of the final models. (a) The relative number of channels left for each convolutional layer with respect to the baseline model. (b) and (c) Comparison of the ROC and precision-recall curves.}
    \label{fig:plot_final}
\end{figure}

We trained the final models at iteration 15 for comparisons. There were 14.72 million parameters for the baseline model, and 0.67 and 1.96 million parameters for the models with and without L1 regularization, respectively. Fig. \ref{fig:plot_final}(a) shows the relative numbers of channels remaining in each layer with respect to the baseline model. We can see that regardless of the use of L1 regularization, more low-level channels were kept in general, which is consistent with the findings that the high-level channels are more problem specific and less transferable \cite{Conference:Yosinski:NIPS2014}. This observation is especially obvious with L1 regularization. \ref{fig:plot_final}(b) and (c) show that even though the number of parameters of the model with L1 regularization was 21 times smaller compared to the baseline, the classification performance was slightly better.

Another important observation is that after 15 iterations, the network trained with L1 regularization only had 969 remaining channels, while still performing better than the baseline VGG network. This is a four fold reduction from the original 4224 channels and allows for faster feature level search and examination of the network.

\begin{figure}[t]
    \centering
    \begin{minipage}[t]{0.2\linewidth}
      \centering
      \includegraphics[width=1\linewidth]{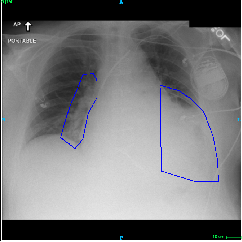}
    \end{minipage}
    \begin{minipage}[t]{0.2\linewidth}
      \centering
      \includegraphics[width=1\linewidth]{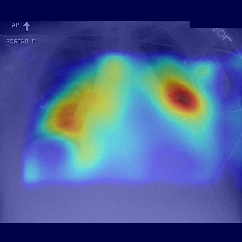}
    \end{minipage}
    \begin{minipage}[t]{0.2\linewidth}
      \centering
      \includegraphics[width=1\linewidth]{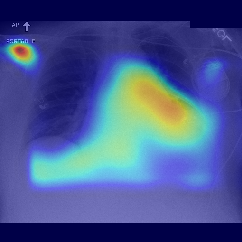}
    \end{minipage}
    \begin{minipage}[t]{0.2\linewidth}
      \centering
      \includegraphics[width=1\linewidth]{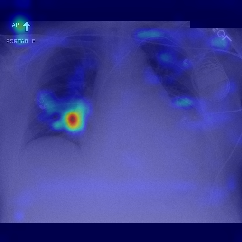}
    \end{minipage}
    \begin{minipage}[t]{0.04\linewidth}
      \centering
      \includegraphics[width=1\linewidth]{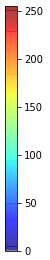}
    \end{minipage}
    \\
    \centering
    \begin{minipage}[t]{0.2\linewidth}
      \centering
      \includegraphics[width=1\linewidth]{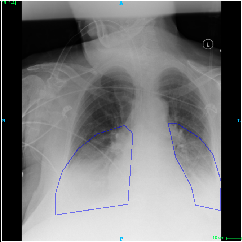}
    \end{minipage}
    \begin{minipage}[t]{0.2\linewidth}
      \centering
      \includegraphics[width=1\linewidth]{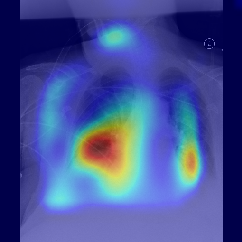}
    \end{minipage}
    \begin{minipage}[t]{0.2\linewidth}
      \centering
      \includegraphics[width=1\linewidth]{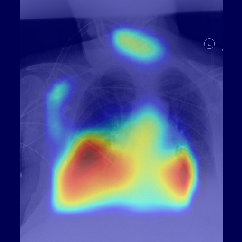}
    \end{minipage}
    \begin{minipage}[t]{0.2\linewidth}
      \centering
      \includegraphics[width=1\linewidth]{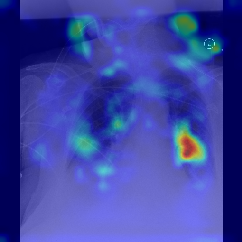}
    \end{minipage}
    \begin{minipage}[t]{0.04\linewidth}
      \centering
      \includegraphics[width=1\linewidth]{colormap}
    \end{minipage}
    \\
    \centering
    \begin{minipage}[t]{0.2\linewidth}
      \centering
      \includegraphics[width=1\linewidth]{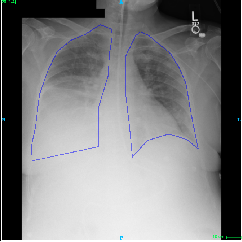}
    \end{minipage}
    \begin{minipage}[t]{0.2\linewidth}
      \centering
      \includegraphics[width=1\linewidth]{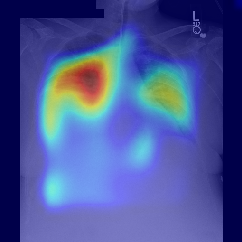}
    \end{minipage}
    \begin{minipage}[t]{0.2\linewidth}
      \centering
      \includegraphics[width=1\linewidth]{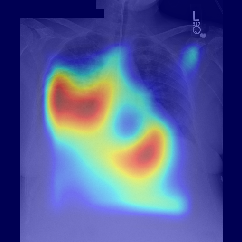}
    \end{minipage}
    \begin{minipage}[t]{0.2\linewidth}
      \centering
      \includegraphics[width=1\linewidth]{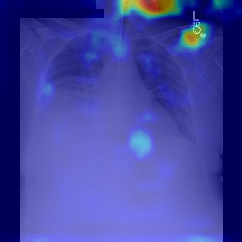}
    \end{minipage}
    \begin{minipage}[t]{0.04\linewidth}
      \centering
      \includegraphics[width=1\linewidth]{colormap}
    \end{minipage}
    \\
    \centering
    \begin{minipage}[t]{0.2\linewidth}
      \centering{Annotation}
    \end{minipage}
    \begin{minipage}[t]{0.2\linewidth}
      \centering{No L1}
    \end{minipage}
    \begin{minipage}[t]{0.2\linewidth}
      \centering{With L1}
    \end{minipage}
    \begin{minipage}[t]{0.2\linewidth}
      \centering{Baseline}
    \end{minipage}
    \begin{minipage}[t]{0.04\linewidth}
      \centering{\hphantom{a}}
    \end{minipage}
    \caption{Examples of class activation maps of positive cases. Left to right: image with annotation in blue, no L1 regularization, with L1 regularization, baseline.}
    \label{fig:cam}
\end{figure}

Fig. \ref{fig:cam} shows examples of class activation maps of positive cases for visual comparisons among the final models. These were produced using the Grad-CAM approach \cite{selvaraju2017grad}. The images were annotated by an expert for the regions of opacity. Although all models correctly classified the cases as positives, the activation maps of the baseline model were barely correlated with the annotated regions, with more irrelevant hot spots outside the lungs. On the other hand, the activation maps of the models with feature selections were better correlated with the annotation. In other words, these smaller networks seems to be also more focused on the correct regions, which is a step towards explainability. The image in the third row shows that the VGG16 activation map has important components outside the lungs, focusing on English characters on the image. Whereas the reduced networks show more activation in marked areas of the lungs.


The experimental results show that using the channel-scaling layers and scaling-and-select strategy, the number of feature channels can be reduced with minimal effects on the classification performance. Furthermore, explainability can also be improved by removing unnecessary feature channels. Using L1 regularization can lead to faster and more network size reduction, but with the tradeoff of the classification performance. Hyperparameters such as the value of the regularization parameter and the threshold value of $s_{li}$ for channel selection can be further adjusted for better tradeoff. Furthermore, because of the relatively few number of trainable network parameters (4,737 at the first iteration), we can use only 20\% of data for training. Instead of keeping all channels from low-level layers and discarding those from high-level layers, we can now select from the best of both worlds using this simple to implement and effective framework.

\section{Conclusion}

We proposed the novel concept of channel-scaling layer to produce smaller and potentially more explainable networks in the transfer learning paradigm for applications in medical imaging. Our method delivered a 95\% reduction in the number of weights, and a four fold reduction in the number of channels starting from a VGG16 network trained on ImageNet, while still maintaining improved classification performance compared to the full network. This large reduction in size can be further amplified if one relaxes the performance requirements. In that scenario, this approach can be used for feature selection, or for building a multi-classifier ensemble that consists of reduced networks each starting from a different pre-trained network.

The black-box nature of deep neural networks is a major obstacle for the widespread use of AI in radiology. As our activation maps show, this concern can be alleviated by our proposed method which seems to focus the network on relevant areas of the image. This along with the reduced number of channels, which facilitates channel level analysis of the network, amounts to an important step towards explainable AI for radiology.

\textbf{Compliance with Ethical Standards:} This research study was conducted retrospectively using human subject data made available in open access by PhysioNet.

\textbf{Acknowledgments:} Authors are employees of IBM Corporation which funded the study, no conflicts of interest.

\bibliographystyle{IEEEbib}
\bibliography{Ref}

\end{document}